\documentclass[letterpaper, 10 pt, conference]{ieeeconf}
\usepackage{siunitx}
\usepackage{times}
\usepackage{amsmath,amssymb}
\usepackage[nolist,nohyperlinks]{acronym}
\usepackage{accents}
\usepackage{graphicx}
\usepackage{bm}
\usepackage{bbm}
\let\labelindent\relax
\usepackage[inline]{enumitem}
\usepackage{array}
\usepackage{multicol}
\usepackage{multirow}
\usepackage{tabulary}
\usepackage{booktabs}
\usepackage{cite}
\usepackage{algpseudocode}
\usepackage[ruled,vlined,linesnumbered]{algorithm2e}
\usepackage[bookmarks=true]{hyperref}
\usepackage{xcolor}
\usepackage[font=small,labelfont=bf]{caption}
\usepackage{subcaption}
\usepackage{amsfonts}
\usepackage{dcolumn}

\newcommand{\code}[1]{{\fontfamily{cmss}\selectfont {#1}}}
\newcolumntype{d}[1]{D..{#1}}

\newcommand{\rv}[1]{\textcolor{black}{{#1}}}
\begin{acronym}
	\acro{CoM}{Center of Mass}
	\acro{DFS}{Depth First Search}
    \acro{RRT}{rapidly exploring random tree}
	\acro{CITO}{Contact-Implicit Trajectory Optimization}
	\acro{MIP}{Mixed-Integer Programming}
	\acro{MIQP}{Mixed-Integer Quadratic Programming}
	\acro{RNN}{Recurrent Neural Network}
	\acro{MLP}{Multilayer Perceptron}
	\acro{MCTS}{Monte Carlo Tree Search}
	\acro{PVMCTS}{Policy-Value Monte Carlo Tree Search}
	\acro{MDP}{Markov-decision Process}
	\acrodefplural{MDP}{Markov-decision Processes}
	\acro{ADMM}{Alternating Direction Method of Multipliers}
	\acro{QP}{Quadratic Program}
    \acro{IL}{imitation learning}
    \acro{BC}{behavior cloning}
    \acro{GCIL}{goal-conditioned imitation learning}
    \acro{GCBC}{goal-conditioned behavior cloning}
    \acro{RL}{reinforcement learning}
    \acro{MDP}{Markov decision process}
	\acro{DoF}{Degree of Freedom}
    \acro{DDPM}{Denoising Diffusion Probabilistic Model}
    \acro{FiLM}{Feature-wise Linear Modulation}
    \acro{FPS}{Farthest Point Sampling}
	\acrodefplural{DoF}{Degrees of Freedom}%
    \acro{GF}{Geometric Fabric}
    \acro{SOCP}{Second-Order Cone Program}
    \acro{PRM}{Probabilistic Roadmap}
\end{acronym}

\IEEEoverridecommandlockouts          
\renewcommand{\baselinestretch}{0.98}

\hypersetup{
    colorlinks=true,
    linkcolor=blue,
    filecolor=green,      
    urlcolor=black,
}
\title{\LARGE \bf
Should We Learn Contact-Rich Manipulation Policies from Sampling-Based Planners?}

\author{Huaijiang Zhu$^{1, 2}$, Tong Zhao$^{2}$, Xinpei Ni$^{2, 3}$, Jiuguang Wang$^{2}$,  Kuan Fang$^{2, 4}$, Ludovic Righetti$^{1,5}$, Tao Pang$^{2}$
 \thanks{$^{1}$New York University}%
  \thanks{$^{2}$Boston Dynamics AI Institute}%
 \thanks{$^{3}$Georgia Institute of Technology}%
 \thanks{$^{4}$Cornell University}%
 \thanks{$^{5}$Artificial and Natural Intelligence Toulouse Institute}%
}

\AddToHook{cmd/@maketitle/after}{\ADDINITIALFIGURE}
\newcommand{\ADDINITIALFIGURE}{%
  \begin{minipage}{\textwidth}
    \expandafter\def\csname @captype\endcsname{figure}%
    \centering
    \includegraphics[width=\textwidth]{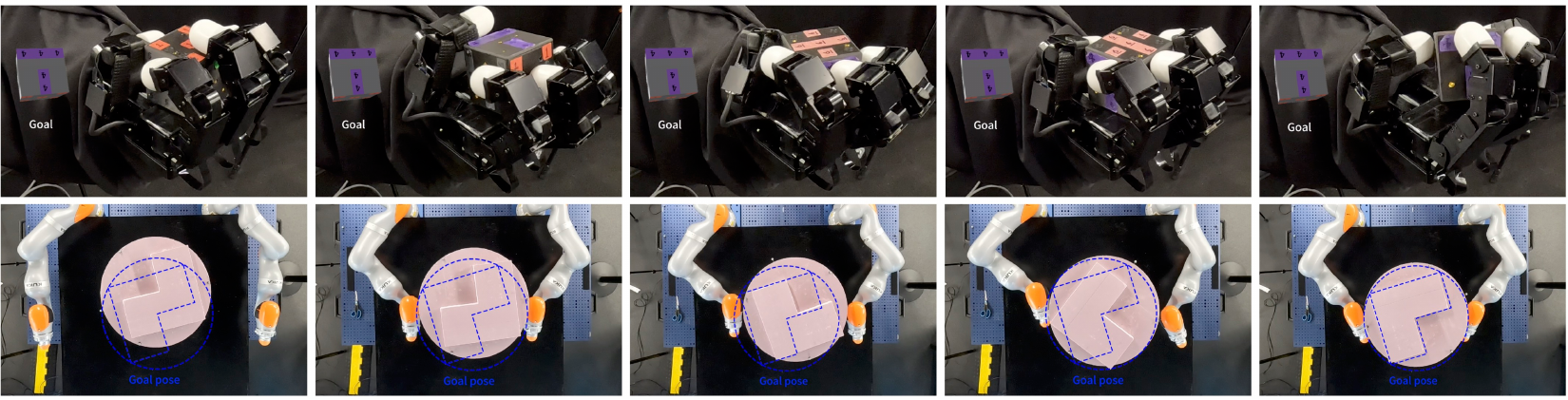}%
    \setcounter{figure}{0}
    \captionof{figure}{
    Contact-rich manipulation tasks considered in this work.
    Top: \textbf{AllegroHand}: In-hand re-orientation with a dexterous hand.
    Bottom: \textbf{IiwaBimanual}: Bimanual manipulation of a cylinder.
    \label{fig:tasks}}
\vspace{-3mm}
  \end{minipage}}

\begin{document}
\maketitle

\begin{abstract}
The tremendous success of behavior cloning (BC) in robotic manipulation has been largely confined to tasks where demonstrations can be effectively collected through human teleoperation. However, demonstrations for contact-rich manipulation tasks that require complex coordination of multiple contacts are difficult to collect due to the limitations of current teleoperation interfaces. We investigate how to leverage model-based planning and optimization to generate training data for contact-rich dexterous manipulation tasks. Our analysis reveals that popular sampling-based planners like rapidly exploring random tree (RRT), while efficient for motion planning, produce demonstrations with unfavorably high entropy. This motivates modifications to our data generation pipeline that prioritizes demonstration consistency while maintaining solution coverage. Combined with a diffusion-based goal-conditioned BC approach, our method enables effective policy learning and zero-shot transfer to hardware for two challenging contact-rich manipulation tasks. Video: \href{https://youtu.be/CxgjJmiiEhI}{\color{magenta} \code{https://youtu.be/CxgjJmiiEhI}}
\end{abstract}

\vspace{-2mm}
\section{Introduction}
\label{sec:intro}
Many everyday manipulation tasks require coordinating multiple contacts with objects using different parts of the body, such as opening a bottle or carrying a large box. To endow robots with true autonomy, acquiring proficiency in these contact-rich dexterous manipulation skills is crucial. However,  executing such skills demands intricate coordination between the hands, the arms, and even the whole body, which leads to a high-dimensional action space. Compared to
single-arm, gripper-based tasks such as pick-and-place, contact-rich dexterous manipulation is also more likely to introduce multi-modality to the solution, i.e., there can be
more than one way to accomplish the task.

Recent years have witnessed a rising trend of learning robotic manipulation skills from human teleoperation~\cite{jang2022bc,shridhar2022cliport,chi2023diffusion,zhao2023learning, zhao2024aloha}. Together with the advances in generative modeling through diffusion models~\cite{ho2020denoising}, \ac{BC} methods have demonstrated their capabilities to learn multi-modal and long-horizon tasks under the simple paradigm of supervised learning~\cite{reuss2023goal, chi2023diffusion}. However, human teleoperation as a data collection method comes with significant limitations. Firstly, as mainstream teleoperation interfaces~\cite{zhao2023learning, chi2024universal} only support tracking the robot end effectors, demonstrations that involve full-arm contacts and multi-finger coordination are challenging to collect. Furthermore, the data collection process is inherently bottlenecked by the availability of human operators, making it difficult to scale robot learning in the same way as vision and language tasks~\cite{kaplan2020scaling}.

\rv{
These limitations have motivated recent work in leveraging synthetic data generated through physics-based simulators. Such data can be produced through various approaches: \ac{RL}, model-based trajectory optimization, or a combination of both. This teacher-student training paradigm, where a \ac{BC} agent learns from an algorithmic expert, has shown success across domains including autonomous driving~\cite{zhang2021end}, legged locomotion~\cite{miki2022learning}, and dexterous manipulation~\cite{chen2023visual}. Given these successes, recent attention has turned to a critical question: how can we produce and curate high-quality data to improve student policy performance? Prior studies define data quality by the distribution shift between the expert that generates the data and the learned policy, arguing that the best data offers sufficient coverage while maintaining low entropy~\cite{belkhale2024data}. }

\rv{
While \ac{RL} methods have achieved impressive progress in contact-rich dexterous manipulation~\cite{chen2023visual,qi2023hand,yin2023rotating,handa2023dextreme}, they do not provide a disentangled way to control the action entropy of the policy. Most \ac{RL} algorithms heavily rely on careful reward shaping, making it difficult to balance exploration, task performance, and constraint satisfaction while controlling data quality. If a reward term were added to encourage low-entropy action, it could also affect exploration and degrade task performance. }

\rv{
In contrast, model-based planning and optimization methods offer more granular control over the data generation process through explicit sampling mechanisms and domain-specific priors. Indeed, recent works have shown that data produced by model-based planning and optimization can be used to directly train an end-to-end policy via \ac{BC} for collision-free motion planning~\cite{dalal2024neural} or legged locomotion~\cite{khadiv2023learning}. For contact-rich manipulation, recent advances in search- and sampling-based planning through contact~\cite{chen2021trajectotree, cheng2023enhancing, pang2023global, zhu2023efficient, graesdal2024towards, suh2025ctr} have emerged as a promising model-based alternative to \ac{RL}. Such approaches can solve contact-rich dexterous manipulation tasks with significantly fewer samples while allowing for straightforward design of cost functions and strict satisfaction of constraints.}

In this work, we study how to leverage model-based planners to produce high-qualit data for learning contact-rich manipulation skills. Our contributions are:
\begin{enumerate}
    \item We show that using inconsistent, high-entropy demonstrations degrades policy performance when learning contact-rich manipulation skills through \ac{BC}. 
    
    \item Drawing from 1), we present a data generation pipeline that produces consistent training data to facilitate effective policy learning.

    \item \rv{We conduct extensive experiments and analysis, validating our hypothesis on challenging contact-rich manipulation tasks using a diffusion-based goal-conditioned \ac{BC} approach.}
\end{enumerate}

In particular, we consider the following two tasks: 
\begin{enumerate}
    \item \textbf{AllegroHand}: an in-hand object rotation task depicted in the top row of Fig.~\ref{fig:tasks}, where a 16-DoF dexterous hand needs to re-orient the cube to a desired orientation.
    \item \textbf{IiwaBimanual}: a bimanual manipulation task depicted in the bottom row of Fig.~\ref{fig:tasks}, where two robot arms are required to rotate an over-sized object by \SI{180}{\degree}.
\end{enumerate} 
Both tasks require reasoning over a long horizon of complex multi-contact interactions with frequent contact switches, presenting significant challenges for both \ac{RL} and human teleoperation.
\vspace{-4mm}

\section{Method}


\begin{figure}[ht]
\centering\includegraphics[width=0.32\textwidth]{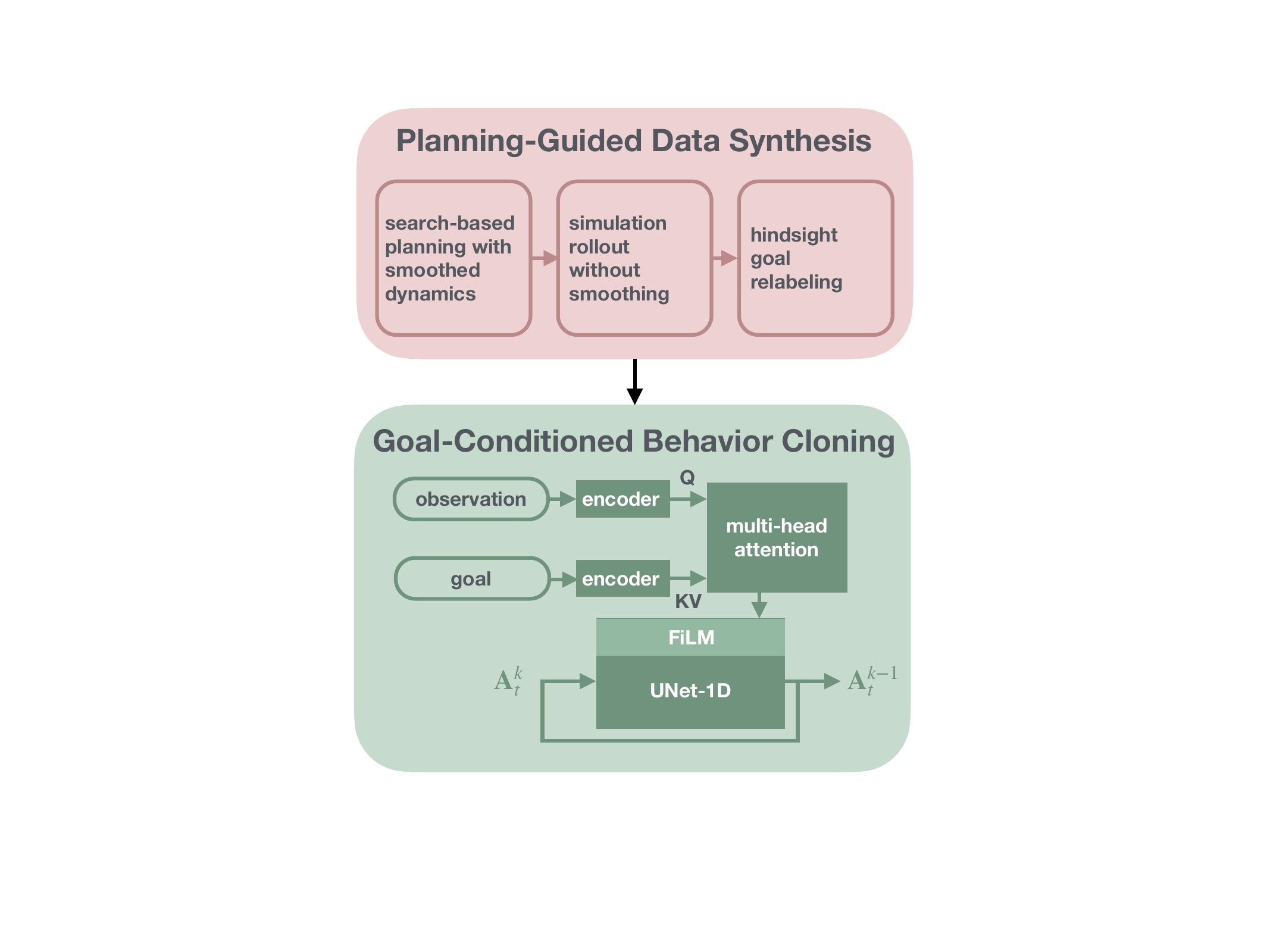}
  \caption{Framework overview.}
  \label{fig:overview}
\end{figure}

Fig.~\ref{fig:overview} provides an overview of our method. First, we obtain a training dataset via a multi-stage data curation pipeline: \begin{enumerate*}
    \item a model-based planner using smoothed contact dynamics~\cite{pang2023global, howell2022dojo, suh2022bundled, suh2022differentiable} proposes a plan,
    \item this plan is then executed in a physics simulator without contact smoothing to produce state-action trajectories.
    \item the reached states are labeled as goals using hindsight goal relabeling.
\end{enumerate*} Using the generated dataset, we learn a goal-conditioned diffusion policy~\cite{chi2023diffusion} mapping the observations directly to actions.
\vspace{-2mm}

\subsection{Planning-Guided Data Synthesis}
\label{sec:planning}
Our method relies on efficient model-based planning. In this section, we first review a \ac{RRT}-based planner proposed in~\cite{pang2023global}, which we summarize in Algorithm~\ref{alg:contact_rrt}. In Section~\ref{sec:entropy}, we present task-specific planner modifications as we show this \ac{RRT}-based planner creates high-entropy data when used for expert demonstration, leading to poor policy performance.

\subsubsection{Planning through contact}
 To allow efficient planning, the planner follows a quasi-dynamic formulation proposed in~\cite{pang2021convex}, where the effect of velocity and acceleration is assumed to be negligible. Hence, the system state $\mathbf{s}\equiv \mathbf{q}$ simply consists of the robot joint positions $\mathbf{q}^{\mathrm{rbt}} \in \mathbb{R}^{n_{\mathrm{rbt}}}$ and the object pose $\mathbf{q}^{\mathrm{obj}}$, either in $\mathbb{SE}(3)$ or $\mathbb{SE}(2)$ depending on the task. The action $\mathbf{a}$ represents the joint position commands that will be tracked by a PD controller. 
\vspace{-3mm}
\begin{algorithm}
\small
\caption{\small{\textbf{Contact RRT}}}\label{alg:contact_rrt}
\textbf{Input:} $\mathbf{q}_0 = [\mathbf{q}^{\mathrm{rbt}}_0, \mathbf{q}^{\mathrm{obj}}_0]\,,\mathbf{q}^{\mathrm{obj}}_\mathrm{goal}\,,p_\mathrm{grasp}$ \\
\textbf{Output:} Tree $\mathcal{T}$ \\
$\mathcal{T}.\texttt{addNode}(\mathbf{q}_0)$ \\
\While{$\mathbf{q}^{\mathrm{obj}} \neq \mathbf{q}^{\mathrm{obj}}_\mathrm{goal}$}{
    \If{$\textsc{UniformRandom}(0, 1) < p_\mathrm{grasp}$}
    {
        $regrasp \gets True$
    }
    \Else{
        $regrasp \gets False$
    }
    \If{$regrasp$}{
    $\mathbf{q}_\mathrm{subgoal}^\mathrm{obj} \gets \mathcal{T}.\texttt{getRandomNode}()$ \\
    $\mathbf{q}_\mathrm{subgoal}^\mathrm{rbt} \gets \textsc{SampleGrasp}(\mathbf{q}_\mathrm{subgoal}^\mathrm{obj})$ \\
    }
    \Else{
    $\mathbf{q}_\mathrm{subgoal}^\mathrm{obj} \gets \textsc{SampleObjectPose}()$ \\
    $\mathbf{q}_\mathrm{subgoal}^\mathrm{rbt} \gets \texttt{None}$ \\
    }
    $\mathbf{q}_\mathrm{nearest} \gets \textsc{Nearest}(\mathbf{q}_\mathrm{subgoal})$ \\
    \If{$regrasp$}{
    $\mathbf{q}_\mathrm{new} \gets \mathbf{q}_\mathrm{subgoal}$
    }
    \Else{
    $\mathbf{q}_\mathrm{new}, \mathbf{a} \gets \textsc{PlanContact}(\mathbf{q}_\mathrm{nearest}, \mathbf{q}_\mathrm{subgoal})$ \\}
    $\mathcal{T}.\texttt{addNode}(\mathbf{q}_\mathrm{new})$ \\
    $\mathcal{T}.\texttt{addEdge}(\mathbf{q}_\mathrm{nearest},  \mathbf{a}, \mathbf{q}_\mathrm{new})$ \\
}
\algorithmicreturn \ $\mathcal{T}$
\end{algorithm}
\vspace{-3mm}
Given the initial robot configuration $\mathbf{q}^{\mathrm{rbt}}_0$, the object pose $\mathbf{q}^{\mathrm{obj}}_0$, and the goal object pose $\mathbf{q}^{\mathrm{obj}}_{\mathrm{goal}}$, the planner switches between sampling a new grasp at the given object pose or sampling an object pose that is reached by solving an inverse dynamics problem. Specifically, given a system state $\mathbf{s} \equiv \mathbf{q} = [\mathbf{q}^{\mathrm{obj}}, \mathbf{q}^{\mathrm{rbt}}]$, $\textsc{PlanContact}(\mathbf{q},\mathbf{q}_\mathrm{des})$ solves for an action $\mathbf{a}$ to bring the object closer to a desired pose $\mathbf{q}^{\mathrm{obj}}_\mathrm{des}$ by solving the following optimization problem
\begin{subequations}\label{eq:id}
\begin{align}
    \underset{\mathbf{a}}{\min} \quad & \frac{1}{2} \lVert \mathbf{q}^{\mathrm{obj}}_{+} - \mathbf{q}^{\mathrm{obj}}_\mathrm{des}\rVert^2 
    \label{eq:id_loss} \\
    \mathrm{s.t.} \quad & \mathbf{q}_{+} = f(\mathbf{q}, \mathbf{a}), \quad g(\mathbf{q}_{+}, \mathbf{a}) \leq \boldsymbol{0}\,,
    \label{eq:id_constraints}
\end{align}
\end{subequations}
where $f(\cdot, \cdot)$ denotes the system dynamics and $g(\cdot, \cdot)$ the state and action bounds such as object pose limits and robot joint limits. The contact dynamics in $f(\cdot, \cdot)$ is smoothed~\cite{pang2023global} and approximated as linear around an appropriately-chosen nominal point and within a convex trust region around the point~\cite{suh2025ctr} such that Problem~\eqref{eq:id} can be solved efficiently using convex optimization. In the case of sampling a new grasp, the planner randomly picks an existing node in the tree and replace the robot configuration with the newly sampled grasp. This new node will be directly added to the tree as we assume the system is always in static equilibrium and the new grasp can be reached by a collision-free planner while the object pose remains unchanged. Once the tree reaches proximity to the goal configuration, it is straightforward to find a shortest path from the root node to the goal.

\subsubsection{Simulation Rollout}
\label{sec:sim_rollout}
Note that we use smoothed contact model in our planner and assume the system to be quasi-dynamic. Such simplifications create discrepancies between the plan and its rollout under second-order dynamics, whether in a full physics simulator or the real world. As such, naively imitating the plan could lead to a policy that deviates from the intended goal. Therefore, we execute the plan in a physics simulator without contact smoothing to obtain the demonstrations. Note that simply executing the entire plan in an open-loop fashion by commanding the planned robot joint angles may lead to a large deviation of the object pose from the plan. We thus rollout the plan in smaller chunks. At the beginning of each chunk, we reset the system state to the planned one. \rv{These chunks naturally arise from the fact that a contact-rich plan can be divided into contact segments and collision-free segments where the robot makes a regrasp and establishes new contacts. A chunk consists of a contact segment and the contiguous collision-free segment. During training, each chunk is treated as a standalone demonstration to avoid discontinuity caused by the state reset.}

\subsection{Goal-Conditioned Behavior Cloning}
We cast the policy learning problem within the framework of \ac{GCIL}~\cite{ding2019goal}. The goal $\mathbf{g} \in \mathbb{SE}(3)$ or $\mathbb{SE}(2)$ is specified by a desired object pose. To address the potential non-Markovianity in our system, we consider a policy that takes as input a history of states $\mathbf{O}_t \equiv \mathbf{s}_{t-h_o:t}$. We also output a sequence of actions $\mathbf{A}_t \equiv \mathbf{a}_{t:t+h_a}$ instead of a single-step action, which has been shown to promote action consistency and reduce compounding errors~\cite{zhao2023learning, chi2023diffusion, zhao2024aloha}. 


We assume access to $N$ demonstrations as trajectories of state-action pairs that are not necessarily optimal
$
    \boldsymbol{\tau}_{i=1}^N = [\mathbf{s}_{1}^i, \mathbf{a}_{1}^i,\mathbf{s}_{2}^i, \mathbf{a}_{2}^i, \cdots, \mathbf{s}_{T}^i]_{i=1}^N\,.
$
To utilize these demonstrations as training data, we use hindsight goal relabeling~\cite{ding2019goal, gupta2019relay, ghosh2019learning}. It hinges on a simple insight: demonstration that falls short of its intended target can nonetheless be viewed as successful for the specific states it did manage to reach. Given an observation history $\mathbf{O}_t$, the action sequence $\mathbf{A}_t$ is valid to reach a future state  $\mathbf{s}_{t+h_a +h_g}$ for any positive integer $0 < h_g \leq T - (t + h_a)$ within the same demonstration. Thus, we can construct a dataset of the tuples
$\mathcal{D} = \cup_{i=1}^N
    \{(\mathbf{O}_t^i, \mathbf{A}_t^i, \mathbf{g}_t^i \equiv \mathbf{s}_{t+h_a+h_g}^i )_{t=h_o}^{T- h_a - h_g}\}\,,$
with $\ 0 < h_g \leq T - (t + h_a)$.
The policy learning problem can thus be formulated as modeling the distribution of the dataset $
    \pi(\mathbf{A} | \mathbf{O}, \mathbf{g}) \equiv p_\mathcal{D}(\mathbf{A}|\mathbf{O}, \mathbf{g})\,.
$

Note that the demonstrations can be multi-modal as the task can be achieved in more than one way. To model such multi-modal data, we choose \ac{DDPM}~\cite{ho2020denoising} to be the action head of the policy, as prior work has demonstrated its ability to capture multi-modal distributions~\cite{reuss2023goal, chi2023diffusion}. At training time, a denoising network $\epsilon_{\boldsymbol{\theta}}(\mathbf{A}_t + \boldsymbol{\epsilon}_k, \mathbf{O}_t, \mathbf{g}, k)$ represented by a 1D \mbox{U-Net}~\cite{ronneberger2015u} learns to predict a Gaussian noise $\boldsymbol{\epsilon}_k$ at different variance levels $k$ from a corrupted sample $\mathbf{A}_t + \boldsymbol{\epsilon}_k$. At inference time, sampling from the learned distribution is achieved by an iterative denoising process
$
    \mathbf{A}_t^{k-1} = \alpha(\mathbf{A}_t^k - \gamma \epsilon_{\boldsymbol{\theta}}(\mathbf{A}_t^k, \mathbf{O}_t, \mathbf{g}, k) + \mathcal{N}(\boldsymbol{0}, \sigma^2 \mathbf{I}))\,,
$
starting from a Gaussian noise $\mathbf{A}_t^K\sim\mathcal{N}(\boldsymbol{0}, \mathbf{I})$ to the sample $\mathbf{A}_t^0$. The denoising network is trained with the loss
$
    MSE(\boldsymbol{\epsilon}_k, \epsilon_{\boldsymbol{\theta}}(\mathbf{A}_t + \boldsymbol{\epsilon}_k, \mathbf{O}_t, \mathbf{g}, k))\,.
$
Following~\cite{chi2023diffusion}, we use \ac{FiLM}~\cite{perez2018film} for observation and goal conditioning. As shown in Fig.~\ref{fig:overview}, the observation history and the goal embeddings are fused by a cross-attention block before being fed to the \ac{FiLM} layer. The encoder is a simple \ac{MLP}.
\section{Data Curation and Planner Modifications}
\label{sec:entropy}
In this section, we take a closer look at how the design choices made to the planning algorithm can significantly impact the policy performance. Recent research~\cite{belkhale2024data} emphasizes the importance of low action entropy in expert demonstrations for \ac{IL}, particularly in low-data regimes. Even with highly expressive models such as diffusion policy, accurately matching the expert distribution becomes challenging when demonstrations have high variability at rarely visited states, as there are insufficient training data to resolve the underlying action distribution. As we will show, despite its widespread success in robot motion planning, \ac{RRT} exhibits this exact unfavorable property when used for generating expert demonstrations. This insight motivates modifications to our planning framework that prioritizes demonstration consistency over planning completeness, which yield data better suited for policy learning.

To illustrate these data generation challenges, we now examine our manipulation tasks and the data curation process in further detail.
\subsection{Bimanual Manipulation}
In the task \textbf{IiwaBimanual}, the manipuland is a cylinder with a height of \SI{0.3}{\meter} and a diameter of \SI{0.6}{\meter}. As this is a planar task, we model the object pose by its position in the $xy$-plane and its yaw angle, i.e. $\mathbf{q}^\mathrm{obj}\equiv[x, y, \theta]$. The task is to rotate the object by \SI{180}{\degree} from the initial orientation of $\theta_0 = \SI{0}{\degree}$  and a random initial position to a fixed goal pose $\mathbf{q}^\mathrm{obj}_\mathrm{goal} = [\SI{0.65}{\meter}, \SI{0}{\meter}, \SI{180}{\degree}]$. We select a large goal orientation of \SI{180}{\degree} to ensure the robots would encounter joint limits during the task, necessitating regrasping to rotate the object to the desired pose.
\paragraph{Random initialization} When generating the demonstrations, we initialize the object position uniformly inside a $\SI{0.4}{\meter}\times\SI{0.7}{\meter}$ region centered at the goal position and sample a random robot configuration that does not collide with the object. Additionally, we remove any demonstrations in which the object pose is outside the pre-defined bounds $\mathbf{q}^\mathrm{obj}_\mathrm{lb} = [\SI{0.35}{\meter}, -\SI{0.35}{\meter}, -\SI{180}{\degree}]\,,\mathbf{q}^\mathrm{obj}_\mathrm{ub} = [\SI{0.85}{\meter}, \SI{0.35}{\meter}, \SI{180}{\degree}]$, as the object will be outside the robot workspace.
\rv{\paragraph{Training parameters} The networks are trained using AdamW~\cite{loshchilov2017decoupled} at a learning rate of $1\times10^{-4}$ for $50$ epochs with a batch size of $256$ samples. We use a cosine learning rate scheduler and adopt exponential moving average of weights to improve training stability as typically practiced for training diffusion models. The number of diffusion denoising steps is set to $100$ during training and $20$ for inference.}

\paragraph{Success criteria} The task is considered successful when the position error is less than \SI{0.1}{\meter} and the orientation error is less than \SI{0.2}{\radian}.

\subsubsection{Planner design}
While the \ac{RRT}-based planner presented in~\ref{sec:planning} efficiently solves contact-rich manipulation tasks such as \textbf{IiwaBimanual}, it is worth noting that the planner samples subgoals at each tree expansion. We hypothesize this sampling strategy leads to a high-entropy action distribution that is more difficult to learn, especially in the low-data regime. To verify our hypothesis, we design a greedy planner as described in Algorithm~\ref{alg:greedy} to generate more consistent demonstrations. This greedy planner iteratively solves Problem~\eqref{eq:id} without sampling subgoals for the object pose. While it still samples the grasp, it only does so when the joint limits are reached.
\vspace{-3mm}
\begin{algorithm}
\small
\caption{\small{\textbf{Greedy Search}}}\label{alg:greedy}
\textbf{Input:} $\mathbf{q}^{\mathrm{rbt}}_0, \mathbf{q}^{\mathrm{obj}}_0, \mathbf{q}^{\mathrm{obj}}_\mathrm{goal}$ \\
\textbf{Output:} Plan $P$ \\
$\mathbf{q}^{\mathrm{rbt}}\gets\mathbf{q}^{\mathrm{rbt}}_0\,, \mathbf{q}^{\mathrm{obj}}\gets\mathbf{q}^{\mathrm{obj}}_0\,, P\gets\texttt{list()}$\\
\While{$\mathbf{q}^{\mathrm{obj}} \neq \mathbf{q}^{\mathrm{obj}}_\mathrm{goal}$}{
    $\mathbf{q}^{\mathrm{rbt}} \gets \textsc{SampleGrasp}(\mathbf{q}^{\mathrm{obj}})$ \\
    \While{$\mathbf{q}^{\mathrm{rbt}}$ not at joint limit and $\mathbf{q}^{\mathrm{obj}} \neq \mathbf{q}^{\mathrm{obj}}_\mathrm{goal}$}{
        $\mathbf{q}, \mathbf{a} \gets \textsc{PlanContact}(\mathbf{q}, \mathbf{q}^{\mathrm{obj}}_\mathrm{goal})$ \\
        $P.$\texttt{append}$(\mathbf{a})$ \\
    }
}
\algorithmicreturn \ $P$
\end{algorithm}
\vspace{-3mm}

\subsubsection{Performance analysis}
To investigate how the action entropy affects the policy performance. We generate datasets of different sizes that respectively contain $100$, $500$, $1000$, and $5000$ demonstrations using the \ac{RRT}-based planner and the greedy planner. Since we do not have access to the distribution $p(\mathbf{a} | \mathbf{O}, \mathbf{g})$, directly calculating its entropy is challenging. Instead, we characterize the action entropy by measuring the action's effect from a few aspects. Specifically, We compute the Shannon entropy $H = -\sum_{b=1}^{B}p_b\log_{B}p_b$
for the following discrete random variables:

\noindent 
\rv{
\textbf{Object velocity direction} We measure the entropy of the discretized object velocity direction. For linear velocity, we divide the $xy$-plane into $B=16$ equal sub-quadrants and assign the linear velocity direction to one of these sub-quadrants. For angular velocity, we assign the movement into three discrete classes: clockwise rotation, counter-clockwise rotation, and no rotation. Moreover, instead of calculating the entropy over the entire state space, we focus only on the $xy$-plane and discretize it into a grid where each cell measures $\SI{0.05}{\meter} \times \SI{0.05}{\meter}$. We obtain the velocity direction by calculating the position and orientation differences between $h_a=60$ steps, where $h_a$ is the length of the action sequences predicted by the policy. The probability $p_b$ is calculated by counting the frequencies of the velocities falling into each sub-quadrants at each cell on the $xy$-plane.}

\noindent 
\rv{\textbf{Progress towards the goal} Additionally, we introduce an interesting quantity to characterize the behavior of the planners. Consider a segment between two consecutive regrasps, we calculate how much progress the planner has made in terms of a weighted distance to the goal $D(q, q_{\mathrm{goal}})= \lVert p_{\mathrm{goal}} - p\rVert + 0.2 \lVert\log({R^{\mathsf{T}}R_{\mathrm{goal}}})\rVert$, where the object pose $q=(p, R)$ is  expressed as the translation vector $p$ and the rotation matrix $R$.
    The progress is then defined by the difference between the initial and the final weighted distance to goal of that segment.}

\noindent 
\rv{\textbf{Regrasp} Finally, a particularly important aspect to consider for contact-rich manipulation skills is the regrasp event, where the robot breaks current contacts and establish new contacts. Intuitively, we prefer regrasps to occur at similar phases and occur less frequently, such that each of them has more samples to learn from, as contact-switch is a much more complex phenomenon than collision-free movement. To visualize the regrasp entropy, we normalize the demonstration completion time to $[0,1]$ and discretize it into $25$ intervals; for each interval, we calculate the entropy for the Bernouli event ``the robot made a regrasp'', hence $B=2$.}

\begin{figure}[t]
\centering
\includegraphics[width=0.48\textwidth]{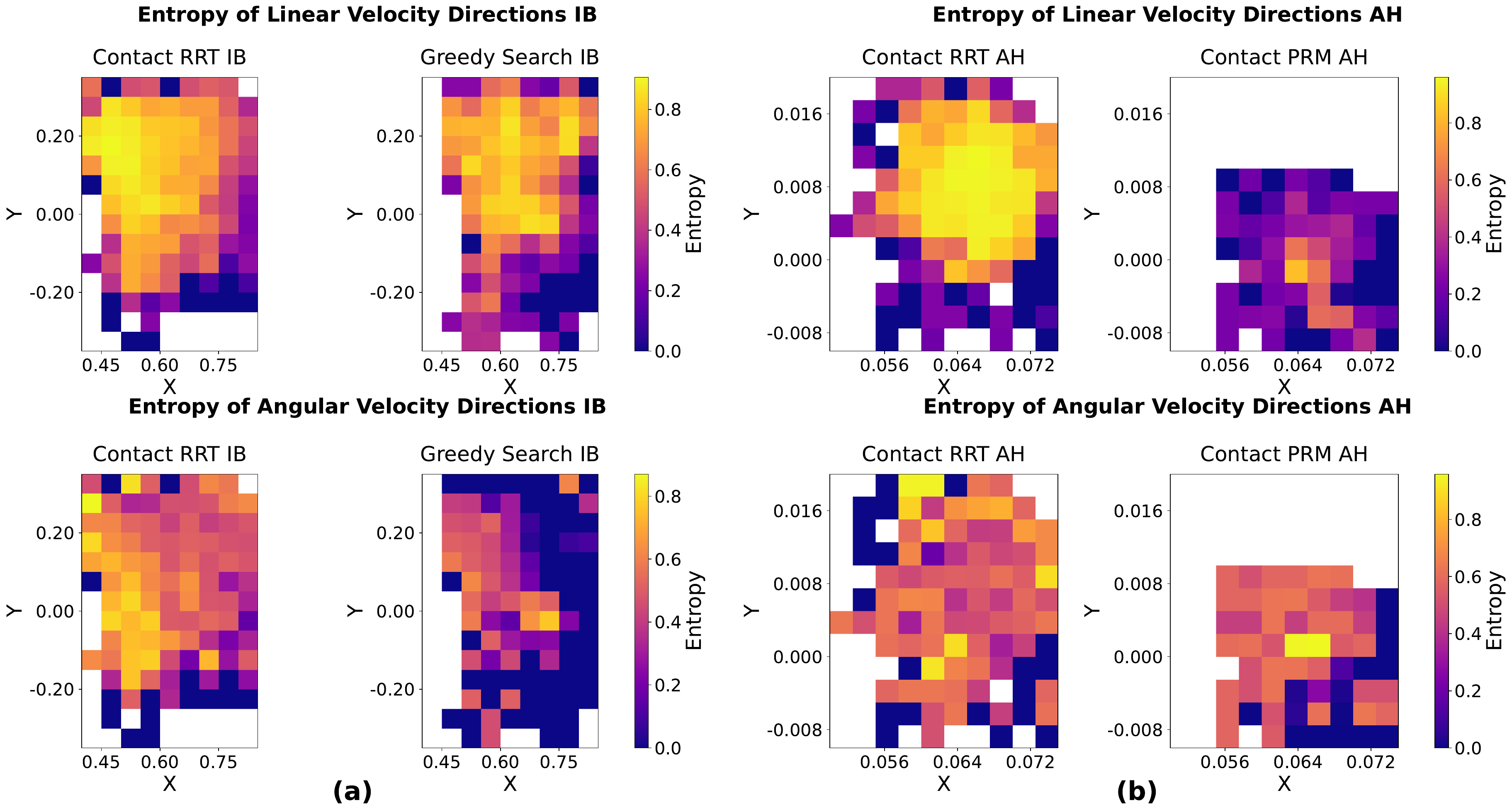}
\caption{Entropy of linear and angular velocity directions of the RRT and greedy datasets for (\textbf{a}) \textbf{IiwaBimaual} (IB) and (\textbf{b}) \textbf{AllegroHand} (AH). White indicates that there is no data.}
\label{fig:entropy}
\end{figure}

\begin{figure}[t]
\centering
\includegraphics[width=0.45\textwidth]{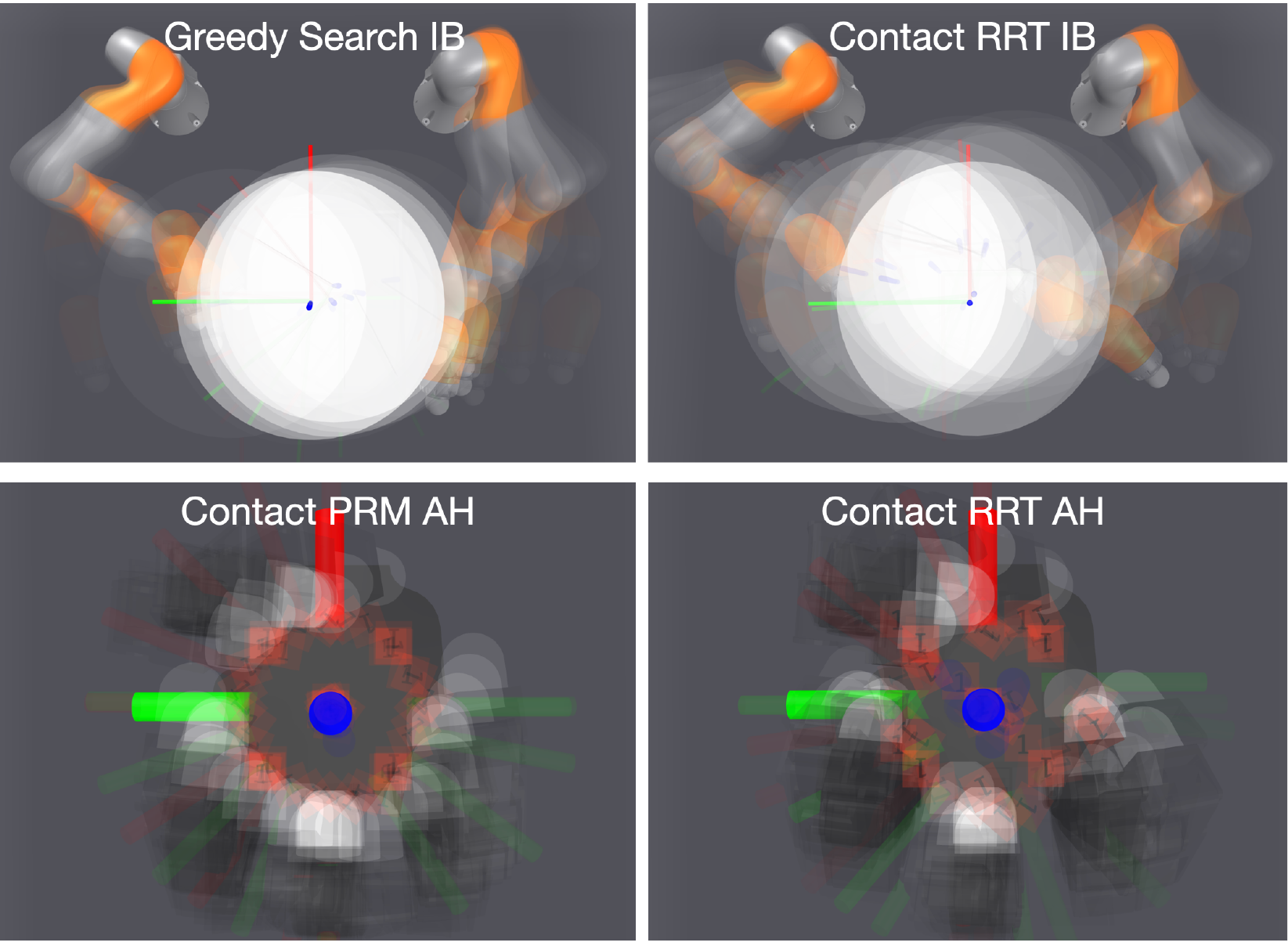}
\caption{Example demonstrations for \textbf{IiwaBimanual }(IB) and \textbf{AllegroHand}(AH). In all subfigures, the solid frames indicate the goal object configuration. For both tasks, contact RRT covers more space by following a more meandering path before reaching the goal than their lower-entropy counterparts.}
\label{fig:example_demos}
\vspace{-0.6cm}
\end{figure}

\begin{figure*}[t]
\centering
\includegraphics[width=0.9\textwidth]{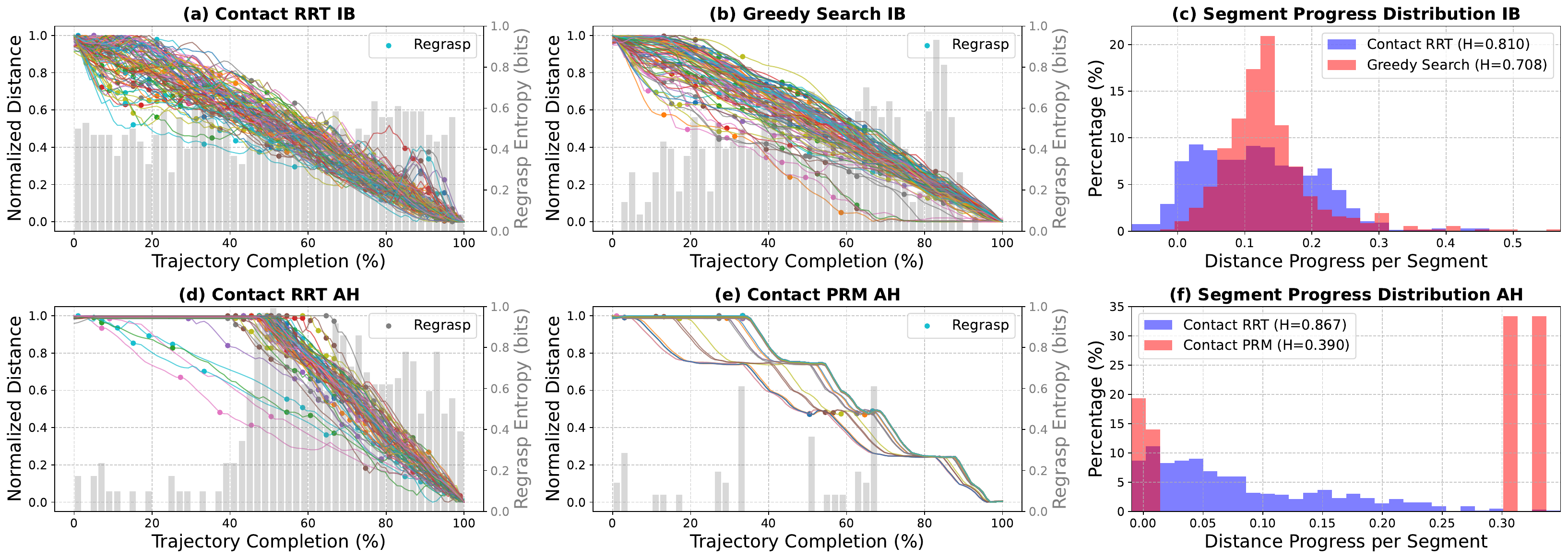}
\caption{
Normalized weighted distance to goal of the object is plotted against trajectory completion percentage for \textbf{(a)} contact RRT IiwaBimanual (IB) and \textbf{(b)} greedy search IB. Each colored curve represents one demonstration trajectory. Dots along a curve represent regrasps. Gray bars represent regrasp entropy for discretized time intervals. \textbf{(c)} Histograms of distance progress per contact segment for IB. 
\textbf{(d)}-\textbf{(f)} Similar plots for the AllegroHand (AH) task. 
We show 100 trajectories for each task to avoid cluttering.
\vspace{-0.4cm}
}
\label{fig:entropy_more}
\end{figure*}

Fig.~\ref{fig:entropy}a shows the plot of the object velocity direction entropy for the \textbf{IiwaBimanual (IB)} task. The \ac{RRT}-based planner has higher entropy in most of the positions than the greedy planner. 
\rv{
Fig.~\ref{fig:entropy_more}c shows the histogram of the per-contact-segment progress towards the goal. 
Typically, a demonstration consists of \qtyrange{5}{10}{} contact segments.
Having a more spread-out distribution, the RRT-based planner exhibits consistently higher entropy. 
Interestingly, for the \textbf{IiwaBimanual} task, the \ac{RRT}-based planner occasionally makes negative progress, moving away from the goal. This is not surprising as \ac{RRT} samples subgoals stochastically. 
Furthermore, the bar plots in Fig.~\ref{fig:entropy_more}a and \ref{fig:entropy_more}b shows that the regrasp entropy of the RRT-based planner stays close to 1 most of the time for the \textbf{IiwaBimanual} task, suggesting that a regrasp can take place with a probability of approximately $50\%$. To further illustrate the difference between the two planners, we show example demonstration trajectories generated by them in Fig.~\ref{fig:example_demos}.
}

Table~\ref{tab:perf_entropy} shows the policy performance measured by the task success rate for $100$ random initial object positions. The randomization range is slightly shrunken to a $\SI{0.3}{\meter}\times\SI{0.6}{\meter}$ region centered at the goal at test time \rv{such that the training set has a broader coverage to handle edge cases that arise at the boundary of the workspace.} Using the same training parameters and network architecture, the policy trained on the data generated by the greedy planner significantly outperforms the one trained on \ac{RRT}-generated data. Indeed, it reaches near perfect success rate given only $100$ demonstrations. The performance gap decreases as we further scale the dataset size, but the policy trained on the \ac{RRT}-generated data plateaus around $85\%$ success rate. Interestingly, the state coverage of the \ac{RRT}-based planner is slightly better than the greedy planner, which is not surprising given its property of probabilistic completeness. However, the performance gap suggests that the relationship between state coverage and policy performance is more nuanced than commonly believed, a finding that aligns with the analysis presented in~\cite{belkhale2024data}.

\vspace{-3mm}

\begin{table}[ht]
\centering
\caption{\textbf{IiwaBimanual} task success rate for different datasets}
\label{tab:gc}
\begin{tabular}{ccccc}
\toprule
\multirow{2}{*}{\textbf{Planner}} & \multicolumn{4}{c}{\textbf{Number of demos}}  \\
& 100 & 500 & 1000 & 5000 \\
\midrule
Contact-RRT & 44\% & 63\% & 88\% & 84\% \\
Greedy Search & 99\% & 98\% & 99\% & 100\% \\
\bottomrule
\end{tabular}
\label{tab:perf_entropy}
\vspace{-0.4cm}
\end{table}

\subsection{In-Hand Re-Orientation}
With the insights gained from the \textbf{IiwaBimanual} task, we now consider a more complex task: 3D in-hand cube reorientation using a 16-DoF Allegro hand. The length of the cube is \SI{6}{\centi\meter}. The object pose $\mathbf{q}^\mathrm{obj} \in \mathbb{SE}(3)$ is represented by the flattened homogeneous transformation matrix with the last row omitted. The goal is specified as the relative transformation between the current and the desired object pose. Two variants of the task are considered: \begin{enumerate}
    \item \textbf{AllegroHand-Easy} where the goal object orientation is constructed by first randomly selecting one of the $24$ cube rotational symmetries (hence the RPY angles are multiples of \SI{90}{\degree}) which we will term as the \emph{canonical} orientations, and then rotating it by a random yaw angle between -\SI{45}{\degree} and \SI{45}{\degree}. This variant is easier because all four corners of the bottom face of the cube are in contact with the palm at the goal pose, which reduces the possibility of slipping.
    \item \textbf{AllegroHand-Hard} where the goal object orientation is uniformly sampled from $\mathbb{SO}(3)$.
\end{enumerate}
In both task variants, the goal position is a predefined nominal position located approximately at the center of the palm.

\subsubsection{Planner modifications} Drawing from our previous analysis, we now consider how to generate more consistent demonstrations with low action entropy for this task. Although the greedy planner performs well for a planar task, the \textbf{AllegroHand} task needs to search through a much higher-dimensional configuration space with more challenging configuration-space obstacles. Therefore, a greedy search strategy will struggle to find a path to the goal. However, exploring the state space in an \ac{RRT} fashion has its own challenges: it not only produces demonstrations with high action entropy as our analysis revealed, but also suffers from inefficiency in high-dimensional space.

As neither \ac{RRT} nor greedy search can generate demonstrations effective for \ac{BC}, we adopt a new global contact planner proposed in \cite{suh2025ctr}. This planner ensures both completeness and consistency by constructing a sparse \ac{PRM} and reusing it for all queries. Specifically, the roadmap includes all canonical orientations as nodes. Furthermore, it can be shown by Monte Carlo estimation that any orientation in $\mathbb{SO}(3)$ lies within \SI{63}{\degree} of a canonical orientation. Hence, given any start and goal orientation, the planner can first find their respective nearest canonical orientations and then traverse through all other canonical orientations to connect the start and goal configuration. The $24$ canonical orientations form a graph that can be connected by three simple primitives \begin{enumerate*}
\item $\texttt{PitchPlus90}$: \SI{90}{\degree} rotation about the world pitch axis,
\item $\texttt{YawPlus45}$: \SI{45}{\degree} rotation about the world yaw axis, and
\item $\texttt{YawMinus45}$: -\SI{45}{\degree} rotation about the world yaw axis.
\end{enumerate*} These primitives are constructed by solving Problem~\eqref{eq:id} iteratively.
The planner also pre-computes a fixed set of grasps for the canonical orientations as opposed to sampling them from all feasible ones. Constraining grasps to a pre-computed set and using fixed primitives reduce the variability inherent in sampling, but still maintains solution diversity through multiple possible shortest paths in the graph. 

In summary, the planner creates a \ac{PRM} where canonical orientations and their associated grasps form nodes, connected by pre-computed primitives. During planning, start and goal configurations are connected to this graph through solving Problem~\eqref{eq:id} and finding the optimal path through the graph using Dijkstra's algorithm. 

\rv{
To compare the behavior of the \ac{RRT}-based planner with the \ac{PRM}-based planner, we design a simplified task \textbf{AllegroHand-Yaw}: rotating the object in-hand by \SI{180}{\degree} along the yaw axis, which is similar to the \textbf{IiwaBimanual} task. We make this simplification because effectively characterizing velocity entropy for objects in $\mathbb{SE}(3)$ is challenging. 
In particular, Fig.~\ref{fig:entropy}b, Fig.~\ref{fig:example_demos} and Fig.~\ref{fig:entropy_more}d-f show similar results to the \textbf{IiwaBimanual} task, suggesting that \ac{RRT}-based planner produces significantly higher-entropy data.}

\subsubsection{Data split}
We collect $1000$ demonstrations of rotating the cube to a uniformly sampled goal orientation in $\mathbb{SO}(3)$. The demonstration always starts from an open-hand configuration with the cube placed at randomly selected canonical orientation, then perturbed by a random translation between $-\SI{3}{\centi\meter}$ and $\SI{3}{\centi\meter}$ and a random yaw rotation between $-\SI{45}{\degree}$ and $\SI{45}{\degree}$. We additionally collect $5000$ demonstrations where the cube is rotated from a canonical orientation by an angle between $\SI{0}{\degree}$ and $\SI{63}{\degree}$ about a uniformly randomly sampled axis, representing the actions required to bring the object to the goal from the nearest canonical orientation. This dataset split addresses the imbalance in our demonstrations. As described previously, the majority of demonstrations consist of pre-computed actions that rotate the cube between canonical orientations. However, the final sequence of actions---rotating the cube from the nearest canonical orientation to the goal---varies significantly. This variable portion requires substantially more training samples to learn effectively.

\subsubsection{Hybrid policy}
\label{sec:hybrid_policy}
To further mitigate the difficulty caused by the dataset imbalance, we implement a hybrid policy approach using two components: \begin{enumerate*}
    \item a main policy trained on $1000$ demonstrations reaching arbitrary goals from perturbed canonical orientations, and
    \item an adjustment policy trained specifically on the $5000$ demonstrations focusing on final orientation adjustments.
\end{enumerate*}
At deployment, when the cube reaches the canonical orientation nearest to the goal, we command the hand to an open-hand configuration and then activate the adjustment policy. In our experiments, this hybrid policy strategy improves success rate by about $10\%$ in simulation compared to a unified policy trained on all $6000$ demonstrations. 

\rv{Interestingly, we note that the training parameters for the \textbf{AllegroHand} tasks are kept the ssame as the ones for \textbf{IiwaBimanual}, suggesting that diffusion policy are not sensitive to hyper parameters tuning, which is consistent with the findings from prior work~\cite{chi2023diffusion}.}
\section{Experiments}
We evaluate the proposed framework in simulation and on hardware to answer the following questions: \begin{enumerate}
    \item Can we learn a policy for contact-rich manipulation from model-based planners?
    \item Can we zero-shot transfer the learned policy to hardware?
\end{enumerate}

\subsection{Evaluation Metric}
We consider the following metrics to evaluate the performance of the policy
\begin{enumerate*}
    \item \textbf{Orientation error}: The orientation error is measured by the difference between the intended and the actual orientation of the object at the terminal step, measured by the norm of the relative axis angle.  
    \item \textbf{Position error} The position error is measured by the $l_2$-norm of the difference between the intended and the actual position of the object at the terminal step.
    \item \textbf{Task success rate} The error threshold for task success is \SI{10}{\centi\meter} in position and \SI{0.2}{\radian} (\SI{11.5}{\degree}) in orientation for \textbf{IiwaBimanual} and \SI{3}{\centi\meter} in position and \SI{0.4}{\radian} (\SI{23.0}{\degree}) in orientation for \textbf{AllegroHand}.
\end{enumerate*}

\subsection{Experiment Setup}
The observation history horizon $h_o$ is $10$ and $3$ steps for the \textbf{AllegroHand} and \textbf{IiwaBimanual} tasks respectively, and the action horizon $h_a$ is $40$ and $60$ steps respectively; for the \textbf{AllegroHand} task, each step takes \SI{0.05}{\second} and for the \textbf{IiwaBimanual} tasks \SI{0.1}{\second}\rv{, which is the same as the discretization step length for the training data}. We choose a relatively long action prediction horizon because we observe jerky motions when the prediction horizon is short, potentially due to the policy switching between different modes of the action distribution. This can be mitigated by warm-starting the inference with the previous action prediction, which we leave for future work.

\begin{table}[ht]
\vspace{-0.3cm}
\setlength{\tabcolsep}{3pt}
\renewcommand{\arraystretch}{1.0}

\begin{subtable}[t]{0.48\textwidth}
\centering
\caption{Task performance in simulation}
\label{tab:sim_results}
\begin{tabular}{cccccc}
\toprule
\multirow{2}{*}{\textbf{Task}} &\textbf{Success}& \multicolumn{2}{c}{\textbf{Position Error} [\SI{}{\centi\meter}]} & \multicolumn{2}{c}{\textbf{Orientation Error} [\SI{}{\degree}] } \\
& \textbf{rate} & Overall & Success & Overall & Success\\
\midrule
AH-EZ-U & 74\% & $1.7 \pm 2.0$ & $1.2 \pm 1.0$ & $31.5 \pm 38.3$ & $13.9 \pm 5.7$\\
AH-HD-U & 57\% & $2.1 \pm 1.1$ & $1.9 \pm 0.8$ & $39.0 \pm 40.1$ & $12.7 \pm 5.7$\\
\midrule
AH-EZ-H & 82\% & $1.3 \pm 0.8$ & $1.1 \pm 0.6$ & $21.8 \pm 18.9$ & $13.7 \pm 5.7$\\
AH-HD-H & 68\% & $1.9 \pm 1.1$ & $1.7 \pm 0.7$ & $28.1 \pm 31.5$ & $12.9 \pm 5.7$\\
\midrule
IB & 99\% & $1.8 \pm 0.5$ & $1.8 \pm 0.5$ & $2.9 \pm 3.1$ & $2.6 \pm 1.4$\\
\bottomrule
\end{tabular}
\end{subtable}


\begin{subtable}[t]{0.48\textwidth}
\centering
\caption{Task performance on hardware}
\label{tab:hw_results}
\begin{tabular}{cccccc}
\toprule
\multirow{2}{*}{\textbf{Task}} &\textbf{Success}& \multicolumn{2}{c}{\textbf{Position Error} [\SI{}{\centi\meter}]} & \multicolumn{2}{c}{\textbf{Orientation Error} [\SI{}{\degree}] } \\
& \textbf{rate} & Overall & Success & Overall & Success\\
\midrule
AH-EZ-H & 62.5\% & $1.5 \pm 1.0$ & $1.0 \pm 0.5$ & $39.8 \pm 40.7$ & $13.7 \pm 5.9$\\
AH-HD-H & 62.5\% & $1.8 \pm 1.0$ & $1.7 \pm 0.9$ & $35.1 \pm 30.3$ & $15.5 \pm 5.2$\\
\midrule
IB & 90\% & $1.9 \pm 0.5$ & $1.9 \pm 0.4$ & $3.5 \pm 5.1$ & $1.8 \pm 0.9$\\
\bottomrule
\end{tabular}
\end{subtable}
\caption{Task performance of the best performing checkpoint in simulation and on hardware.
AH-EZ-U stands for AllegroHand-Easy-Unified; AH-HD-H stands for AllegroHand-Hard-Hybrid; IB stands for IiwaBimanual.
For AH tasks, only the hybrid policy is evaluated on hardware.}
\label{tab:exp_results}
\vspace{-0.6cm}
\end{table}

\subsection{Simulation Evaluation}
We evaluate the tasks in simulation using the best performing checkpoint during training. We execute the policy from $100$ random initial object poses and report the success rate along with the error mean and standard deviation in Table~\ref{tab:sim_results}. Error metrics are calculated in two ways: across all trials, and separately for successful trials only, as some failure cases (e.g., when the object falls off the table or the hand) might result in large errors, and the statistics could be skewed by these outliers.

\paragraph{\emph{\textbf{AllegroHand}}} As described in Section~\ref{sec:hybrid_policy}, we adopt a hybrid policy strategy for the \textbf{AllegroHand} task, as we observe it improves the success rate for both \textbf{AllegroHand-Easy} and \textbf{AllegroHand-Hard} compared to the unified policy trained on all demonstrations. We believe this performance gap is caused by the data imbalance in our dataset. While we overweight the fine adjustment demonstrations ($5000$ vs. $1000$) in our dataset when training the unified policy, determining the optimal data mixture ratio remains a complex challenge that exceeds the scope of this work.  For both unified and hybrid policies, one of the most common failure modes we observe is the policy fails to react to out-of-distribution scenarios not present in training data. This is unsurprising given our use of pre-computed primitives in the demonstrations. While techniques like DAgger~\cite{ross2011reduction} could potentially address these failures through data augmentation with corrective behaviors, our planner is currently unable to find solutions from arbitrary system configurations, making it difficult to apply DAgger. 
\paragraph{\emph{\textbf{IiwaBimanual}}}The policy for \textbf{IiwaBimanual} is trained on $100$ demonstrations generated by the greedy planner. While the policy has high success rate in simulation, we do occasionally see chattering-like behaviors where the policy switches between different action modes, hence clockwise and counter-clockwise rotations. As a result, the policy often takes longer than necessary to complete the task.

\subsection{Hardware Evaluation}
For hardware experiments, we use an OptiTrack motion capture system to provide the object pose. Table.~\ref{tab:hw_results} shows the error metrics for the hardware experiments.
\paragraph{\emph{\textbf{AllegroHand}}}
For hardware evaluations of the \textbf{AllegroHand} task, we adopt the hybrid policy strategy. We place the cube at the center of the palm with an initial orientation close to the identity at the beginning of each evaluation. To make sure the goals are spread out across $\mathbb{SO}(3)$, we generate goal orientations by applying random rotational perturbations to the $24$ canonical orientations. For \textbf{AllegroHand-Easy}, we add a random yaw rotation within the range $[-\SI{45}{\degree}, \SI{45}{\degree}]$. For \textbf{AllegroHand-Hard}, we sample the perturbation using the axis-angle representation, where the rotation angle ranges from $\SI{0}{\degree}$ to $\SI{63}{\degree}$ about a random 3D unit vector axis (recall that any element in $\mathbb{SO}(3)$ can be reached this way). For both task variants, $15$ out of $24$ trials are successful, representing a $62.5\%$ success rate. Hence, the success rate for \textbf{AllegroHand-Hard} is comparable with simulation while \textbf{AllegroHand-Easy} sees some performance degradation. The most common failure mode occurs when the object lands in configurations not present in training data, leaving the policy unable to recover. While this failure mode exists in simulation, it occurs more frequently on hardware, likely due to the sim-to-real gap. Interestingly, we observe the policy sometimes taking a long action sequence to reach canonical orientations that could have been achieved with fewer primitives. We hypothesize this is due to the network incorrectly interpolating between goals in the training data.
\paragraph{\emph{\textbf{IiwaBimanual}}} We execute the policy with the object placed at $20$ initial positions; $18$ out of $20$ trials are successful, representing a $90\%$ success rate. The failure cases occur when the initial object position is placed at the boundary of the training data distribution, and the orientation error for the failed trials are around \SI{15}{\degree}, only slightly above the error threshold.
We note that our real-world setup has a slight model mismatch from the simulation. For example, the mass of the object is \SI{1.25}{\kilo\gram}, but in simulation, we set it to be \SI{1.0}{\kilo\gram}; the object shape is not perfectly cylindrical and measures only \SI{0.59}{\meter} in diameter instead of the \SI{0.6}{\meter} in simulation. We further note that we do not domain randomize parameters such as the geometry of the object and the robot or the friction coefficients during training or data generation.

\section{Conclusion}
In this work, we demonstrate that model-based motion planning offers a compelling alternative to human teleoperation for generating training data for contact-rich manipulation tasks. This approach eliminates the bottleneck of manual data collection while enabling the generation of demonstrations for complex tasks that are challenging to demonstrate through current teleoperation interfaces, such as those involving full-arm contacts and multi-finger coordination.

However, our analysis reveals an important nuance: the effectiveness of learning from planned demonstrations heavily depends on how we design the planning algorithm. While popular sampling-based planners like \ac{RRT} excel at global planning, they can generate demonstrations with high action entropy that are difficult to learn from, especially in low-data regimes. This insight motivates us to modify our data generation pipeline to prioritize demonstration consistency while maintaining adequate state space coverage and solution diversity. Our empirical results show that policies trained on more consistent demonstrations significantly outperform those trained on data from standard \ac{RRT} planners.
By combining careful planner design with diffusion-based generative modeling, our approach successfully learns challenging contact-rich manipulation skills that can be zero-shot transferred to hardware. These results suggest that model-based planning is indeed a valuable tool for scaling up \ac{BC} beyond simple gripper-based tasks. \rv{Nevertheless, we acknowledge that generating data entirely from simulation comes with its own limitations. For example, contact interactions between non-rigid objects cannot yet be realistically simulated or effectively planned, making it difficult to apply our approach to soft hands or deformable objects. This may be addressed by using simulated data for pre-training and real-world data for post-training, which we leave for future work.}

\bibliographystyle{IEEEtran}
\bibliography{root}
\end{document}